\UseRawInputEncoding 
\documentclass[conference]{IEEEtran}
\IEEEoverridecommandlockouts
\usepackage{cite}
\usepackage{amsmath,amssymb,amsfonts}
\usepackage{algorithmic}
\usepackage{graphicx}
\usepackage{textcomp}
\usepackage{xcolor}
\def\BibTeX{{\rm B\kern-.05em{\sc i\kern-.025em b}\kern-.08em
    T\kern-.1667em\lower.7ex\hbox{E}\kern-.125emX}}
\begin{document}

\title{DREAM Lite:\\Simplifying Robot Assisted Therapy for ASD\\
\normalsize{\medskip Development, deployment and validation of the DREAM European project''s exploitation part}
\thanks{This work has been funded by the European FP7 project ``DREAM'' (Grant Agreement Number 611391), and was written for and presented during the Social Robots in Therapy and Care workshhop, part of the ACM/IEEE International Conference on Human Robot Interaction, Daegu, 2019}
}

\author{\IEEEauthorblockN{Alexandre Mazel}
\IEEEauthorblockA{\textit{Innovation Department} \\
\textit{SoftBank Robotics Europe}\\
Paris, France \\
amazel@softbankrobotics.com}
\and
\IEEEauthorblockN{Silviu Matu}
\IEEEauthorblockA{\textit{Department of Clinical Psychology and Psychotherapy} \\
\textit{Babes-Bolyai University}\\
Cluj-Napoca, Romania \\
silviu.matu@ubbcluj.ro}
}
\maketitle

\begin{abstract}
Robot-Assisted Therapy (RAT) has successfully been used to improve social skills in children with autism spectrum disorders (ASD). The DREAM project explores how to deliver effective RAT interventions to ASD children within a supervised-autonomy framework for controlling the robotic agent, which could decrease the burden on the clinicians delivering such interventions. In this paper we describe how to use in real life settings the experimental protocols that were developed and extensively tested in the DREAM Project, as well as their deployment and validation in a new ecological study conducted by clinicians.
\end{abstract}

\section{Introduction}
Autism Spectrum Disorder (ASD) is characterized by impairments in social interactions and communication, usually accompanied by restricted interests and repetitive behaviors \cite{b1}.
Previous research has shown that the integration of robotic agents with psychological interventions for ASD patients could facilitate the delivery of such interventions, especially for children, as robotic agents might act as mediators between human models and ASD children, see \cite{b2,b3,b4}. 
The EC-FP7 funded DREAM project \cite{b5} (Development of Robot-Enhanced therapy for children with AutisM spectrum disorders) aims to investigate the clinical utility of Robot-Assisted Therapy (RAT) delivered within a supervised-autonomy framework. This project targets to increase the robot's autonomy during RAT in order to reduce the workload of the therapist, psychologist or teacher, by letting parts of the intervention be taken over by the robot, without altering the clinical utility of the intervention. The RAT protocols that were developed based on this approach where extensively tested in a large scale randomized clinical trial.

In order to develop a semi-autonomous solution for delivering RAT, a complex sensing and control system had to be put in place, including multiple sensors (i.e., high-definition cameras, depth cameras, microphones). Such sensors were needed to collect, measure and understand the different reactions of ASD children, so that the robotic agent can respond in a contingent way, following the treatment rationale. As the DREAM project is near completion, we had to think about a simplified system that would allow other therapists to reproduce independently the RAT protocols developed in the project, without the use of such a complex technological setup. The goal of this paper is to describe this simplified system and the results of its clinical validation using a large sample of heterogeneous children with ASD symptoms coming form multiple educational centers.

\section{The DREAM Project}
One of the most important questions and challenges of the development and delivery of RAT solutions for ASD children, are related to the robot's control and autonomy. A common approach described in the literature is to use a Wizard of Oz setup, where a human, hidden or not, remotely controls every behaviour of the robot, including what it says. This method does require an additional human agent that controls the robot, but which also needs to understand the clinical intervention that is being delivered with the help of the robot. These requirements reduce the probability for large scale adoption of RAT in educational and recovery institutions, as they increase the personnel costs and the level of expertise required to conduct such interventions. On the other hand, a fully autonomous robot that conducts a therapeutic intervention is not conceivable with the current technology, but it also might raise serious ethical questions, as well as trigger negative reactions from parents and the general public. The goal of the DREAM Project is to develop a hybrid solution, one in which the robot is sufficiently autonomous not to be a burden for the therapist, but at the same time under the control of the therapist, in order to deliver the therapy effectively.

The DREAM project is build based on the integration of five main components. (1) The first component is a clinical one, build by psychotherapists, and consists in the development and validation of the intervention protocols that are delivered during RAT sessions. The RAT protocols developed in the project have been tested using a rigorous large scale randomized clinical trial aiming to see if they are at least as efficacious as standard interventions. These intervention protocols target key skills that have been shown to relate to the deficits that are characteristic to ASD children: imitation, turn taking, and joint attention. (2) The second component of the project is of a system for recording the child's behavioral responses (e.g. movements and actions), called sensory analysis. This perception component, comprising an array of high-resolution video cameras,  depth sensors and microphones, allows the system to map the behavior of the child and compare it to an expected pattern that is consistent with a good performance. This automatic interpretation of child behaviors has great clinical value, as it reduces the effort required from the therapist to keep track of the progress of the child while conducting the intervention, or to review the video recordings in order to gather data about performance. (3) The third component consists of a decision and response sub-system that uses the input coming from sensory analysis, which is contextually interpreted based on the task that is being performed, and controls the behavior of the robot so that it will responds in a way that is in line with the therapeutic requirements. To facilitate such a complex decision, this sub-system learns form the operator's feedback (a trained psychologist who supervises and corrects if necessary the interpretations made by the decision sub-system) during therapy session. (4) The main goal of the fourth component of the project is developing an ethical framework for research and practical applications of this technology.  (5) Finally, the fifth component of the project is dealing with how to disseminate and to harness the social and economic values of the RAT intervention hat has been developed and tested extensively using  a simplified implementation that it can be easily deployed in hospitals and educational institutions. It is this part that we will describe in the next paragraphs. The DREAM project has been presented in more details in other publications \cite{b6}.

\section{Exploitation Solution}
The deployment of a solution that has been developed as part of a research project can pose several challenges. First, the solution has to comprise as many of the features that have led to the experimental positive results, or otherwise it might not keep all of the advantages that were demonstrated via research. On the other hand, when deploying, several constraints coming from the ecosystem must be taken into account, such as ease of implementation, reduced time for setup and deployment, robustness and economic constraints. The exploitation phase of the DREAM project, called DREAM Lite, focused mainly on simplifying two aspects: removing sensors external to the robot and merging the role of the assistant (controlling the robot) and the therapist. The decision to remove the external sensors greatly reduced the decision-making capabilities of the robot, but it was considered as not feasible otherwise. Next, the use of a specifically developed tablet interface allows the therapist to control the robot while keeping enough attention for the interaction with the child, without necessarily needing an assistant. The robot chosen for this exploitation is the SoftBank's NAO robot\footnote{NAO the humanoid robot by SoftBank Robotics, https://www.softbankrobotics.com/emea/en/nao}, the same robot that was initially used in developing and testing the clinical protocols. Moreover, its design and robustness make it a good tool to deploy the RAT intervention in ecological settings.

\begin{figure}[htbp]
\centerline{\includegraphics[width=55mm]{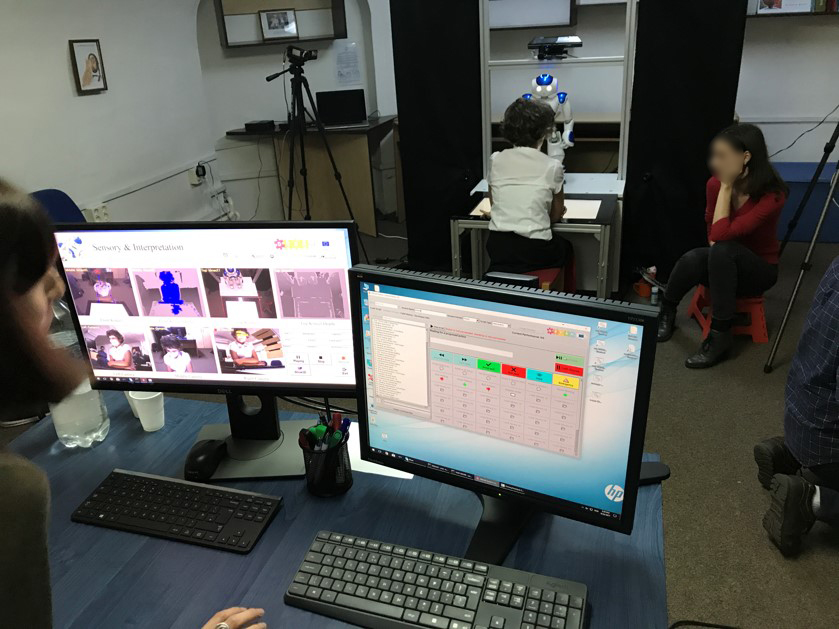}}
\caption{The DREAM project experimental setup.}
\label{fig1}
\end{figure}

\begin{figure}[htbp]
\centerline{\includegraphics[width=55mm]{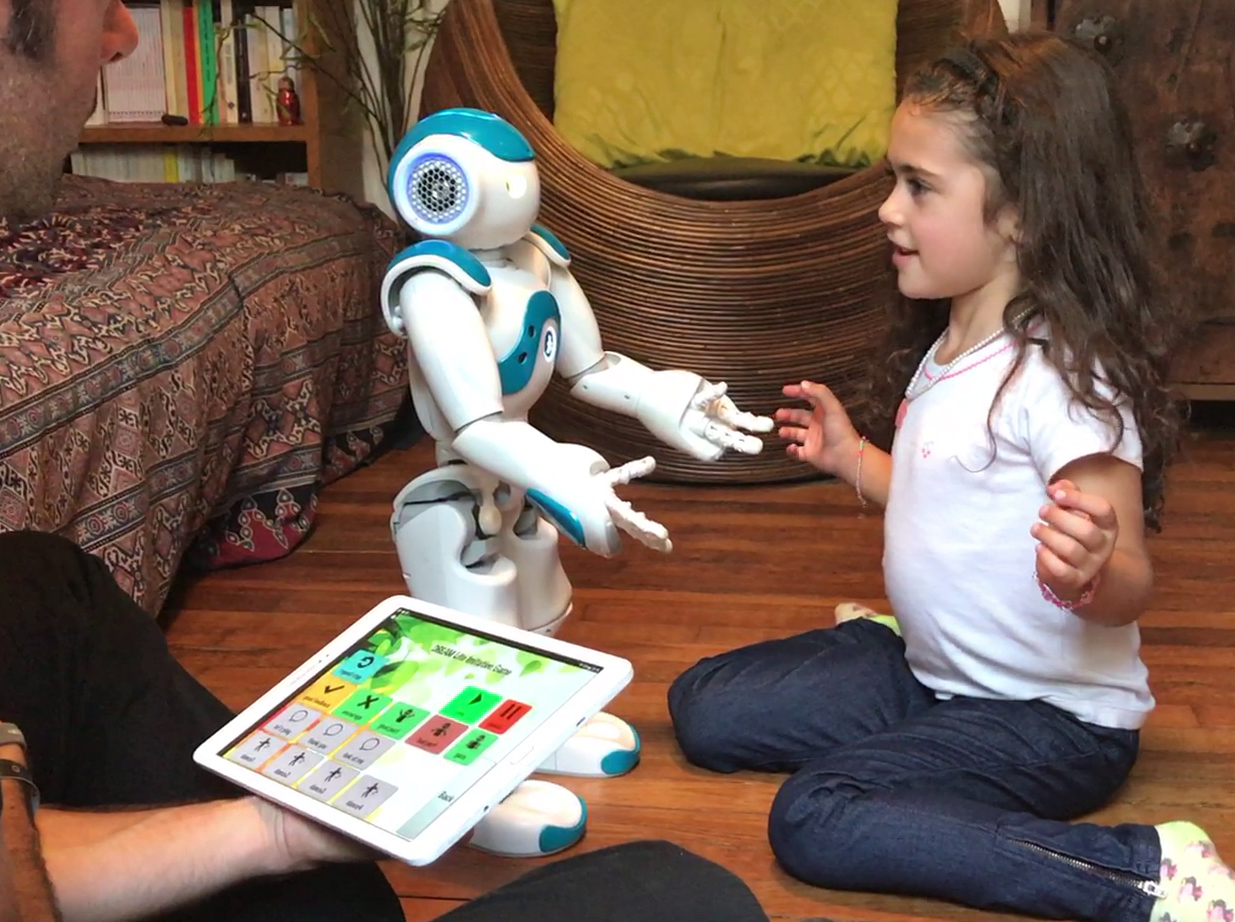}}
\caption{The DREAM Lite solution being tested.}
\label{fig2}
\end{figure}

The figure~\ref{fig1} was taken during the DREAM project experimentation, and shows the installation of cameras and other sensors, as well as the roles of the therapist and that of the assistant who controls the robot. This gives an overview of the framework that had to be put in place: cameras positions must be calibrated and the complete system is likely to be unfamiliar to a child. In comparison, the figure~\ref{fig2} shows the first tests of the DREAM Lite solution. In addition to its simplicity, it gives an idea of the ecological aspect of such a solution: the child and the robot interact in a more familiar environment and the therapist has a great proximity with him or her. The child and the therapist can also move during the experiment if necessary.

\section{Applications descriptions}
\begin{figure}[htbp]
\centerline
{%
\setlength{\fboxsep}{0pt}%
\setlength{\fboxrule}{0.1pt}%
\fbox{{\includegraphics[width=52mm,height=31mm]{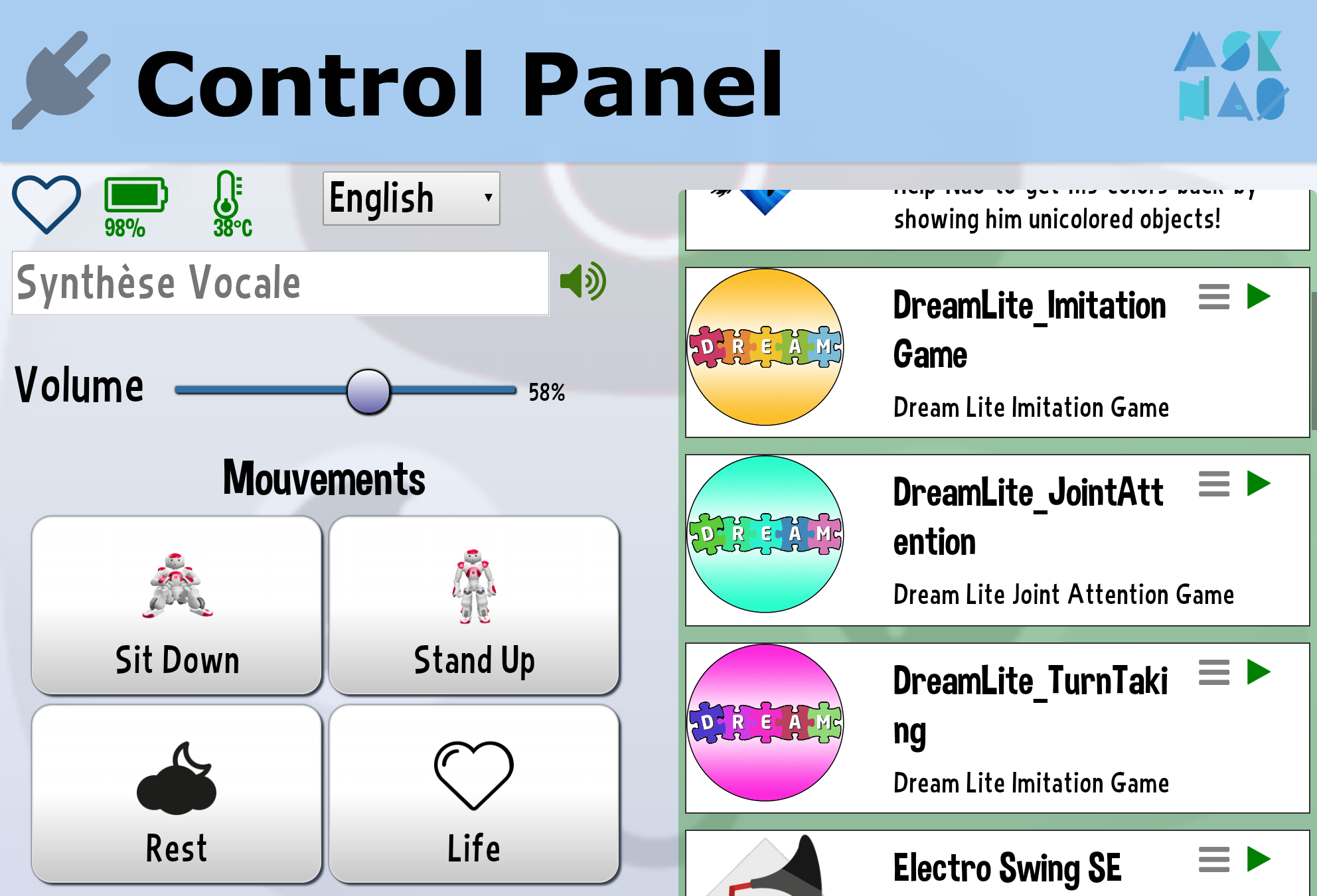}}}
}%
\caption{The AskNAO Tablet Menu.}
\label{fig3}
\end{figure}

The applications developed within the framework of this exploitation were based on the ``AskNAO Tablet'' interface, distributed by ERM\footnote{ERM Robotique, http://www.erm-robotique.com/.} in France; the launcher of this interface is depicted in figure~\ref{fig3}. The applications that can be run using the tablet interface cover each protocol developed and tested in the DREAM project: imitation, turn taking, and joint attention. Particular attention was paid during applications' development to make them very easy to use, while allowing responsiveness, so that the robot can adapt as quickly as possible to the needs of the child.

\subsection{Imitation}
\begin{figure}[htbp]
\centerline
{%
\setlength{\fboxsep}{0pt}%
\setlength{\fboxrule}{0.1pt}%
\fbox{{\includegraphics[width=58mm]{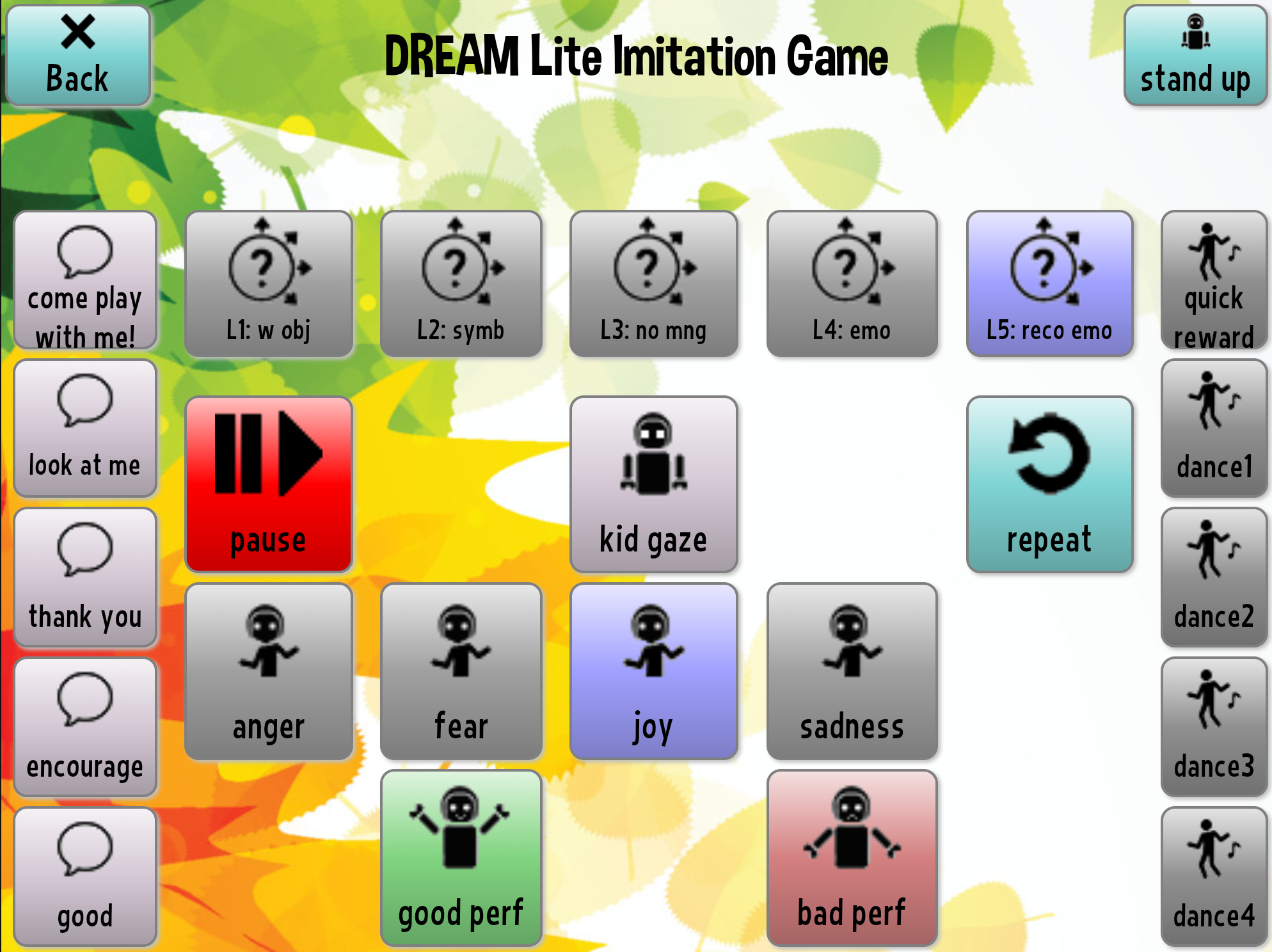}}}
}%
\caption{The Imitation App Tablet Interface.}
\label{fig4}
\end{figure}

The purpose of this application is to improve the children ability to imitate. This application includes five different sets of exercises: (1) a level in which the child must imitate movements and sounds related to a specific object, such as moving a toy car while also imitating the sound of the engine; (2) a symbolic set of exercises where the child must do the same type of movements as the previous ones, but this time with a generic purpose object, such as a wood play block; (3) the next level requires the child to imitate movements that represent daily life gestures, such as waving hand while saying goodbye;  (4) the interface depicted in figure~\ref{fig4} is used for the next set of exercises which are related to the imitation of movements representing emotional expressions (anger, fear, joy and sadness); (5) finally, an additional set of exercises was added and targets the recognition of emotional expressions.

\subsection{Joint Attention}
This application makes use of one of the features of the humanoid robot NAO, namely its ability to simulate human gaze. In this exercises, the child learns to engage in joint attention episodes, by looking at objects designated by the robot, which are placed either on the left side or the right side of the robotic agent. Depending on the level of difficulty, the robot will use multiple cues to show the object of interest (pointing with its arm, using a vocal indication, and simulate a gaze by turning its head towards the object), or it will use just a single cue (simulate a gaze). This task teaches children to synchronize their attention focus with that of the play partner.

\subsection{Turn Taking}
This application teaches the child to play a structured game with a partner while respecting the rules of the game: the partners make their moves in turns and one has to wait for his or her turn. In this exercises the child and the robot must take turns in indicating their favorite food or activity by choosing one of several cards placed in front of them (each card belongs to a category such as sweets, dishes, or hobbies).

\subsection{NAO Blind}
\begin{figure}[htbp]
\centerline{\includegraphics[width=55mm]{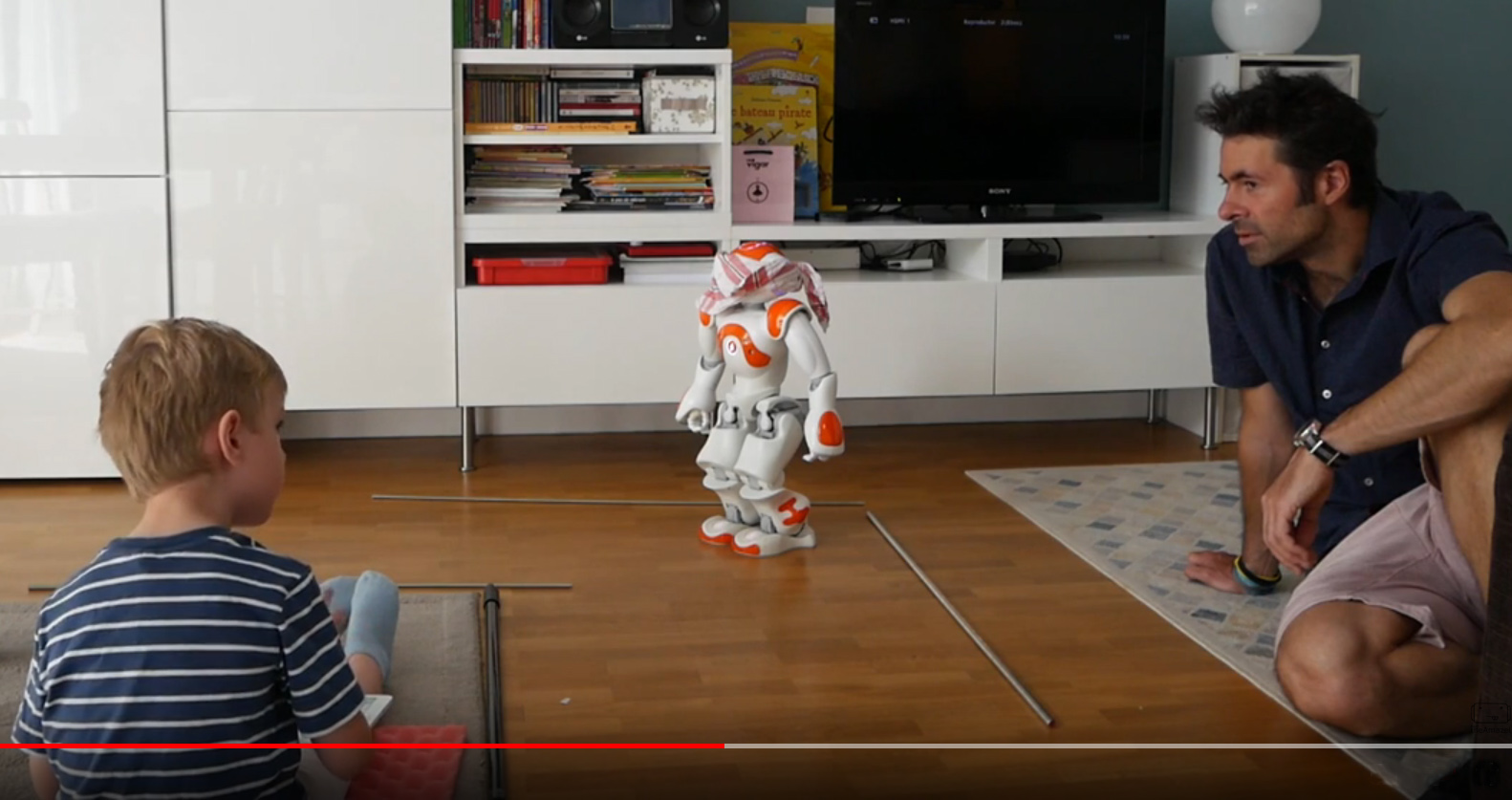}}
\caption{NAO Blind first testing.}
\label{fig5}
\end{figure}

The setup used during the clinical experimentation conducted in the DREAM project did not allow for the exploitation of  all the advantages brought by a mobile humanoid robot such as NAO. Namely, the robot remained mostly static during exercises, which limited the physical interactions between the child and the robot, and reduced the dynamics of the game play. The overcome this issues, we have developed an extra application to explore the potential provided by the mobility that the robot is capable of. This gave rise to a fourth application in which the robot simulates that it has been blinded and asks the help of the child to walk and move in the environment. The child can guide the robot by controls on the tablet, and thus the application offers opportunities for interactions that are both dynamic and close. An example of setup can be see in figure ~\ref{fig5}.

\section{Methodology}
In order to test the DREAM Lite solution and to validate its feasibility, stability and its real-life clinical utility, we used an ecological study design - an effectiveness study - which was implemented in 10 educational institution in Cluj-Napoca, Romania. The sample consisted of children with ages between 3 and 10 years old,  who had an ASD diagnosis or had ASD symptoms as presented in their medical file. Sixty eight children were chosen for the study to date. Children were included if they were in the specified age range, they had a formal diagnosis or psychological assessment indicating the presence of ASD symptoms, and they had the minimum abilities to understand and follow the assessment or intervention tasks. We did not exclude children based on the presence of other mental health and neurological problems (e.g., learning disorders).  The children were randomly allocated to one of two conditions: a) a RAT intervention delivered via the DREAM Lite solution, or b) a wait-list (WL) condition in which the children waited for two weeks before they receive the same intervention. All children were recruited from special education institutions and thus children allocated to the control condition (WL) followed regular education activities for two weeks. The comparisons between  RAT and WL actually reflects the benefits that the children could gain form the DREAM Lite solution if this therapeutic tool would be added to their treatment routine.
The intervention consisted of five sessions: a behavioral assessment based on the Autism Diagnostic Observation Schedule-2 (ADOS-2) \cite{b7}, which was meant to evaluate the baseline of the children' social interaction skills; three intervention sessions of about 30 minutes each, performed daily or every second day, which targeted imitation, joint attention, and turn-taking skills; and a final evaluation session which was identical to the initial one. The intervention sessions followed the discrete trial training \cite{b8} in which the robot partner presented a discriminative stimulus in the form of an instruction (e.g., ``do like me'', for the imitation intervention), waited for the behavioral response of the child, and offered a contingent feedback (positive social reinforcement or an indication to correct the behavior). The WL group was also assessed at the beginning of the wait period and just before starting the RAT intervention. 

\section{Measurements}
We measured the social interaction skills of the children based on the dedicated module form the ADOS-2, as well as the imitation, joint attention and turn-taking skills, based on behavioral tasks performed in interaction with a human partner. We also measured the performance in imitation, joint attention and turn-taking during each intervention session. To increase measurement accuracy, an external observer coded the performance during these sessions using a standardized observation grid. The performance of the child in each trial, across each tasks and each level of difficulty was coded as good or bad performance. We also video recorded each sessions and an experienced supervisor reviewed the initial coding. Finally, we used self-report questionnaires to assess parents' satisfaction with the intervention, as well as the perception by the staff in the institutions regarding the usability and acceptability of the DREAM Lite solution.

\section{Conclusion}
Experiments are undergoing and the first results will be available starting with March 8th, 2019. Based on the results of the clinical trial conducted during the project and the positive results of RAT for ASD children in literature, we expect that the DREAM Lite will have a positive impact on the social skills of the children. Also, given the simplified interface and the ease of use, we expect to have positive responses from staff and parents related to its utility. The next steps will largely depend on the results of the study. Yet, several key points will have to be carefully analyzed and weighted: 1) the degree to which delivering the intervention using the DREAM Lite solution can replicate the results from the clinical trial; 2) the level of engagement that the solution is able to elicit in this heterogeneous population (sub-groups analysis based on age, gender, severity of symptoms,  and other relevant variables will be considered); 3) the usability of the tablet-based solution and the attitudes of the clinicians and teachers towards adopting it.

Also, given that most of the children were affected by multiple developmental disorders (e.g., language disorders, attention deficit and hyperactivity disorder), it was decided not to include the NAO Blind application in this experimentation, as it might have increased the complexity of the therapy sessions. Thus, a next step would be to test this application, either as a stand alone RAT activity, or in combination with the other DREAM Lite applications, using a sample that is capable of following more complex instructions.

RAT interventions for ASD children have greatly evolved over the last decade, and have moved from simple interaction studies, showing that these children are engaging in interactions with the robot partner, to clinical studies, investigating their potential in making significant changes in the lives of this vulnerable population. The DREAM Lite solution however, moves the field one step forward and offers an intervention tool that has been previously clinically tested and validated, and which can be easily implemented in real-life. DREAM Lite represent one of the first examples of disseminating evidence-based RAT interventions for ASD.

\section*{Contributions}
AM and SM contributed equally to this manuscript.

\section*{Acknowledgment}
This works was supported by the project ``Development of Robot-Enhanced therapy for children with AutisM spectrum disorders'', co-financed by the the European Commission, grant agreement no. 611391, https://www.dream2020.eu.

\vspace{12pt} 
\begin{flushright}v1.0b-Arxiv\end{flushright}
\end{document}